\documentclass[letterpaper, 10 pt, conference]{ieeeconf}

\IEEEoverridecommandlockouts                              

\usepackage{times}

\usepackage{graphicx}
\graphicspath{ {./figures/} }
\usepackage{subcaption}
\usepackage{xcolor}
\usepackage{pifont}
\usepackage{xcolor,pifont}
\usepackage[utf8]{inputenc}
\usepackage{hyperref}
\usepackage{float}
\usepackage{listings}
\usepackage{algorithm}
\usepackage{algpseudocode}

\usepackage{cellspace}
\definecolor{codegreen}{rgb}{0,0.6,0}
\definecolor{codegray}{rgb}{0.5,0.5,0.5}
\definecolor{codepurple}{rgb}{0.58,0,0.82}
\definecolor{backcolour}{rgb}{0.95,0.95,0.92}
\lstdefinestyle{mystyle}{
    backgroundcolor=\color{backcolour},   
    commentstyle=\color{codegreen},
    keywordstyle=\color{magenta},
    numberstyle=\tiny\color{codegray},
    stringstyle=\color{codepurple},
    basicstyle=\ttfamily\footnotesize,
    breakatwhitespace=false,         
    breaklines=true,                 
    captionpos=b,                    
    keepspaces=true,                 
    numbers=left,                    
    numbersep=5pt,                  
    showspaces=false,                
    showstringspaces=false,
    showtabs=false,                  
    tabsize=2
}
\definecolor{green(ryb)}{rgb}{0.4, 0.69, 0.2}
\definecolor{red(ryb)}{rgb}{1.0, 0.15, 0.07}
\newcommand{\cmark}{\textcolor{green(ryb)}{\ding{51}}}%
\newcommand{\xmark}{\textcolor{red(ryb)}{\ding{55}}}%

\usepackage{array} 
\newcolumntype{M}[1]{>{\centering\arraybackslash}m{#1}} 
\usepackage{threeparttable} 

\title{Aerostack2: A Software Framework for Developing Multi-robot Aerial Systems}

\author
{Miguel Fernandez-Cortizas$^{1}$, Martin Molina$^{1,2}$, Pedro Arias-Perez$^{1}$, \\Rafael Perez-Segui$^{1}$, David Perez-Saura$^{1}$, Pascual Campoy$^{1}$
\thanks{$^{1}$
Computer Vision and Aerial Robotics Group (CVAR), Centre for Automation and Robotics (C.A.R.), Universidad Politécnica de Madrid (UPM-CSIC), 28006 Madrid, Spain. Correspondence e-mail:  miguel.fernandez.cortizas@upm.es}
\thanks{$^{2}$ Department of Artificial Intelligence, Universidad Politécnica de Madrid (UPM), 28040 Madrid, Spain}
\thanks{This work has been supported by the project COPILOT ref. Y2020\textbackslash EMT6368, funded by Madrid Government under the R\&D Synergic Projects Program. We acknowledge the support of the European Union through the Horizon Europe Project No. 101070254 CORESENSE. This work has also been supported by the project INSERTION ref. ID2021-127648OBC32 and the project RATEC ref: PDC2022-133643-C22 funded by the Spanish Ministry of Science and Innovation. The work of the third author is supported by the Grant FPU20/07198 of the Spanish Ministry for Universities. The work of the fifth author is supported by the Spanish Ministry of Science and Innovation under its Program for Technical Assistants PTA2021-020671. }}

\begin{document} 


\baselineskip24pt


\maketitle



\normalsize 

\color{black}
\begin{abstract}

The development of autonomous aerial systems, particularly for multi-robot configurations, is a complex challenge requiring multidisciplinary expertise. Unlike ground robotics, aerial robotics has seen limited standardization, leading to fragmented development efforts. To address this gap, we introduce Aerostack2, a comprehensive, open-source ROS 2 based framework designed for creating versatile and robust multi-robot aerial systems. Aerostack2 features platform independence, a modular plugin architecture, and behavior-based mission control, enabling easy customization and integration across various platforms. In this paper, we detail the full architecture of Aerostack2, which has been tested with several platforms in both simulation and real flights. We demonstrate its effectiveness through multiple validation scenarios, highlighting its potential to accelerate innovation and enhance collaboration in the aerial robotics community.
\newline
\footnotesize
\newline
Source code: \url{https://github.com/aerostack2/aerostack2}\\
Documentation: 
\url{https://aerostack2.github.io/}

\end{abstract}
\section{Introduction}





Developing autonomous aerial systems from scratch is a challenging task that requires extensive expertise in many different areas, such as aerodynamics, control systems, sensor integration, or AI algorithms. This is a common problem in the robotics field, so in recent years the robotics community has witnessed the development of several software stacks focused on the control and guidance of ground robots and articulated robots. Navigation2 \cite{nav2} and MoveIt \cite{moveit} are two examples that have gained widespread adoption. However, the same level of collaboration and standardization has not been observed in the field of aerial robotics. Even when frameworks for aerial systems have been developed, they often have a narrow focus, like low-level control or working on a concrete platform, which limits their usefulness in broader applications. This fragmentation of effort can make it difficult to take advantage of the strengths of each framework in a particular application, where relying on well-tested and robust algorithms is critical to enabling developers to focus on innovation and customization rather than software engineering.

To address these challenges, this paper proposes a collaborative framework for aerial robotics that brings users and developers together to work toward a common goal. We present Aerostack2, a ROS 2-based framework designed to facilitate the development of complex autonomous aerial systems. By building a solid architecture that supports versatile and robust aerial systems, Aerostack2 aims to enhance research in aerial robotics and accelerate its application in various industrial domains. This paper presents an overview of our proposed framework and discusses its potential impact in the field.

The Aerostack2 software framework presented represents a new iteration designed from scratch and built on the foundation laid by its predecessor, Aerostack \cite{sanchez2016aerostack}. Aerostack2 has been developed by learning from the strengths and weaknesses of its predecessor, which has been used successfully in our research lab for over six years, contributing not only to academic research, but also to industrial projects and international robotics competitions such as IARC2014 \cite{sanchez2015IARC},  IMAV2016 \cite{sampedro2017fully} or MBZIRC 2020 \cite{rodriguez2021autonomous}.


\subsection*{Contributions}


In this work, we present Aerostack2 a novel framework for designing Aerial Robotics systems in an easy and versatile way. The main contributions of our work, compared with other solutions are:

\begin{itemize}
    \item \textbf{Platform Independence}: Offers platform-agnostic design, facilitating seamless integration across various drones and simulators, enhancing adaptability and interoperability.
    
    
    \item \textbf{Behavior-based mission definition and monitoring}: Implements behavior-based logic for mission control, providing a structured approach for task formulation and execution monitoring.
    
    \item \textbf{Plugin oriented architecture:} Enables dynamic reconfiguration and easy integration of new components, enhancing system adaptability and long-term maintainability.
    
    \item \textbf{Versatility}: Serves as a versatile framework supporting a modular architecture that allows users to easily customize or replace each module, providing flexibility for different application scenarios.
    
    \item \textbf{Validation in different showcases}: These contributions are validated in a variety of environments, including controlled laboratory settings and industrial simulations, demonstrating their reliability and suitability for real-world use.
\end{itemize}

\section{Related Work}

In order to control an unmanned aerial vehicle (UAV) autonomously, there are two complementary parts: the Flight Control Unit (FCU), systems that focus on low-level control of the aircraft, allowing it to fly stably, and the high-level control frameworks that are responsible for providing further autonomy to the vehicle, making decisions, and commanding the FCU to perform a specific task. Below, we will discuss these two parts of the state of the art in more detail.

\subsection{\textbf{Programmable Flight Control Unit}}

FCUs can be classified into two blocks: generic or embedded, each presenting distinct advantages and limitations. Generic FCUs offer versatility, accommodating various frames and components for customizable configurations, which is beneficial for developers. However, their integration and calibration require technical expertise and time. Embedded FCUs, integrated into complete aerial platforms, provide a user-friendly experience with pre-calibrated hardware for reliable performance. This simplicity suits users seeking ready-to-fly solutions, but may limit adaptability to new technologies or specific needs.

\begin{table}[h]
\begin{threeparttable}
\centering
\footnotesize
\begin{tabular}{|l|c|c|c|}
    \hline
    \textbf{Flight Controller} & \textbf{Open Source} & \textbf{Simulation} & \textbf{Generic} \\
    \hline
    \hline
    Pixhawk/PX4 & OSH, OSS & SITL, HITL & \cmark \\
    \hline
    Ardupilot & OSS & SITL, HITL & \cmark \\
    \hline
    Paparazzi & OSH, OSS & SITL, HITL & \cmark \\
    \hline
    Crazyflie & OSH, OSS & SITL, HITL & \xmark \\
    \hline
    DJI Matrice & Propietary & HITL & \xmark \\
    \hline
    Parrot & Propietary & SITL & \xmark \\
    \hline
    Skydio & Propietary & - & \xmark \\
    \hline
\end{tabular}
\begin{tablenotes}
    \item OSH: Open Source Hardware, OSS: Open Source Software, SITL: Software In The Loop, HITL: Hardware In The Loop.
\end{tablenotes}
\end{threeparttable}
\caption{Comparison of relevant programmable flight systems.}
\label{tab:flight-controllers}
\vspace{-0.3cm}
\end{table}

In 2018 Ebeid et al. presented a survey of open-source hardware and software comparing their main features \cite{ebeid2018survey}. In Table \ref{tab:flight-controllers} some relevant flight controller projects are listed.  These projects may cover both hardware and software development of these controllers. They range from Open Source Hardware (OSH) and Open Source Software (OSS) to proprietary commercial controllers. Furthermore, most of them allow to simulate their behavior in a fully simulated way or Software-in-the-Loop (SITL) or in a Hardware-in-the-Loop (HITL) way, where the autopilot code actually runs on the specific hardware of the controller. This makes it possible to improve the validation of the systems created, leading to more robust and reliable solutions.

Although some flight controllers can be programmed to perform simple autonomous tasks, such as waypoint navigation and basic obstacle avoidance, they lack the capabilities required to handle complex tasks. That is why they are combined with high-level controllers that improve the robustness and autonomy of these vehicles, allowing them to perform complex autonomous missions.


\subsection{\textbf{High-Level Frameworks for Aerial Robotics}}

\begin{table*}
\begin{threeparttable}
\centering
\footnotesize
\begin{tabular}{ | m{2.7cm} | M{1.2cm} | M{1.3cm} | M{1.3cm} | M{1.1cm} | M{1.3cm} | M{1.1cm} | M{1.1cm} | M{1.1cm} | M{1.1cm} | M{1.1cm} | }
    \hline
    \textbf{Framework} & \textbf{Modular} & \textbf{Maintained} & \textbf{Tested in} & \textbf{Middle-ware} & \textbf{Industrial use} & \textbf{Multi-frame} & \textbf{Multi-agent} & \textbf{Multi-platform} & \textbf{Plugin oriented} \\
    \hline
    \hline
    Aerostack \cite{sanchez2016aerostack} & \cmark & \xmark & S,RL,RO & ROS & \cmark & \xmark &  \cmark & \cmark & \xmark \\
    \hline
    Agilicius \cite{foehn2022agilicious} & \cmark & \cmark & S,RL,RO & ROS & \xmark & \cmark &  \xmark & \xmark & \xmark \\
    \hline
    CrazyChoir \cite{pichierri2023crazychoir} & \xmark & \cmark & S,RL & ROS 2 & \xmark & \xmark & \cmark & \xmark & \xmark \\
    \hline
    CrazySwarm2 \cite{preiss2017crazyswarm} & \cmark & \cmark & S,RL & ROS 2 & \xmark & \xmark & \cmark & \xmark & \cmark \\
    \hline
    GAAS \cite{gaas} & \cmark & \xmark & S & ROS & - & \xmark & \xmark & \xmark & \xmark \\
    \hline
    KumarRobotics \cite{mohta2018fast} & \xmark & \xmark & S,RL,RO & ROS & \cmark & \xmark & \xmark & \cmark & \xmark \\
    \hline
    MRS UAV System \cite{baca2021mrs} & \cmark & \cmark & S,RL,RO & ROS & \cmark & \cmark & \cmark & \cmark & \xmark \\
    \hline
    RotorS \cite{Furrer2016} & \cmark & \xmark & S & ROS & - & \xmark & \xmark & \xmark & \xmark \\
    \hline
    UAL \cite{real_ijars20} & \xmark & \xmark & S,RL,RO & ROS & \cmark & \cmark & \xmark & \cmark & \xmark \\
    \hline
    XTDrone \cite{xiao2020xtdrone} & \cmark & \cmark & S & ROS & - & \xmark &  \xmark & \xmark & \xmark \\
    \hline
    \hline
    \textbf{Aerostack2 (Ours)} & \cmark & \cmark & S,RL,RO & ROS 2 & \cmark & \cmark &  \cmark & \cmark & \cmark \\
    \hline
\end{tabular}
\begin{tablenotes}
    \item S: Simulation; RL: real experiments in the laboratory; RO: real experiments outside the laboratory.
\end{tablenotes}
\end{threeparttable}
\caption{Comparison of relevant open-sourced software stacks for aerial robotic systems. Industrial use is only evaluated for the frameworks that support real flights.}
\label{tab_aerial_stacks}
\vspace{-0.3cm}
\end{table*}

Among general purpose frameworks, we can find some solutions, such as Aerostack \cite{sanchez2016aerostack}, a ROS-based framework that integrates various computational solutions such as computer vision and motion planning, developed through extensive use in our research lab. The MRS UAV system \cite{baca2021mrs} is another ROS-based system designed for onboard use with PX4 and DJI compatible controllers, suitable for both indoor and outdoor applications, and designed for multi-robot experiments. The UAV Abstraction Layer (UAL) \cite{real_ijars20} further standardizes the interfaces of UAVs on different autopilots, simplifying higher-level algorithm development for both simulated and real platforms. 

For more specialized applications, Agilicious \cite{foehn2022agilicious}, is a framework optimized for agile quadrotor flight. It features onboard vision sensors, GPU-accelerated hardware, and a versatile software stack for real-time perception and control. Similarly, the KumarRobotics flight stack \cite{mohta2018fast} focuses on autonomous navigation in cluttered, GPS-denied environments, performing all sensing and computation onboard without human intervention after launch.

In the realm of simulation and experimentation, CrazyChoir \cite{pichierri2023crazychoir} offers a ROS 2 toolbox for working with swarms of Crazyflie nano-quadrotors, enabling complex distributed tasks in both simulations and real-world experiments. Similarly, Crazyswarm2 \cite{preiss2017crazyswarm}, is a toolbox designed for a large swarm of Crazyflies flying in dense indoor formations, obtaining accurate flights for up to 50 drones. XTDrone \cite{xiao2020xtdrone} provides a comprehensive simulation platform for various types of UAV, facilitating the transition from algorithm testing in simulations to deployment in real UAVs. RotorS \cite{Furrer2016} is another modular MAV simulation framework, designed for quick research integration and easy transition from simulation to real MAV applications.

Finally, the Generalized Autonomy Aviation System (GAAS) \cite{gaas} presents a simulation platform aimed at fully autonomous VTOL and multirotors. It uses lidar, HD map relocalization, and path planning for robust autonomous flight, including human-carrying applications, and integrates with national air traffic control systems.


Table \ref{tab_aerial_stacks} compares some relevant characteristics of these open-source frameworks. The table shows that, while most frameworks are modular, only half are actively maintained. A limited number support ROS 2 as middleware, and even fewer are designed for industrial use. Approximately half of the frameworks support multi-agent operations, but multi-frame and multi-platform capabilities are less common. Very few frameworks feature a plugin-oriented design, which allows for easy addition of new components and supports dynamic reconfiguration of the system. Furthermore, testing environments vary, with some frameworks tested only in simulations, while others have been validated in real laboratories and field conditions.



Aerostack2 has been designed to cover all these mentioned features, allowing for a wide variety of use cases. It supports multi-agent and multi-platform operations, making it highly adaptable. Its plugin oriented architecture facilitates easy addition of new components and dynamic reconfiguration, making it easy to maintain and suitable for both research and industrial applications.
\section{A Stack of Software Components for Aerial Robotics}

Aerostack2 framework is organized in the form of a software stack with components distributed in different hierarchical layers, as can be seen in Fig. \ref{fig:software_stack}. Components in one layer control the components of the layers below (corresponding to less complex functionalities). The layers are as follows:

\begin{figure*}[htb!]
\centering
\includegraphics[width=0.85\textwidth]{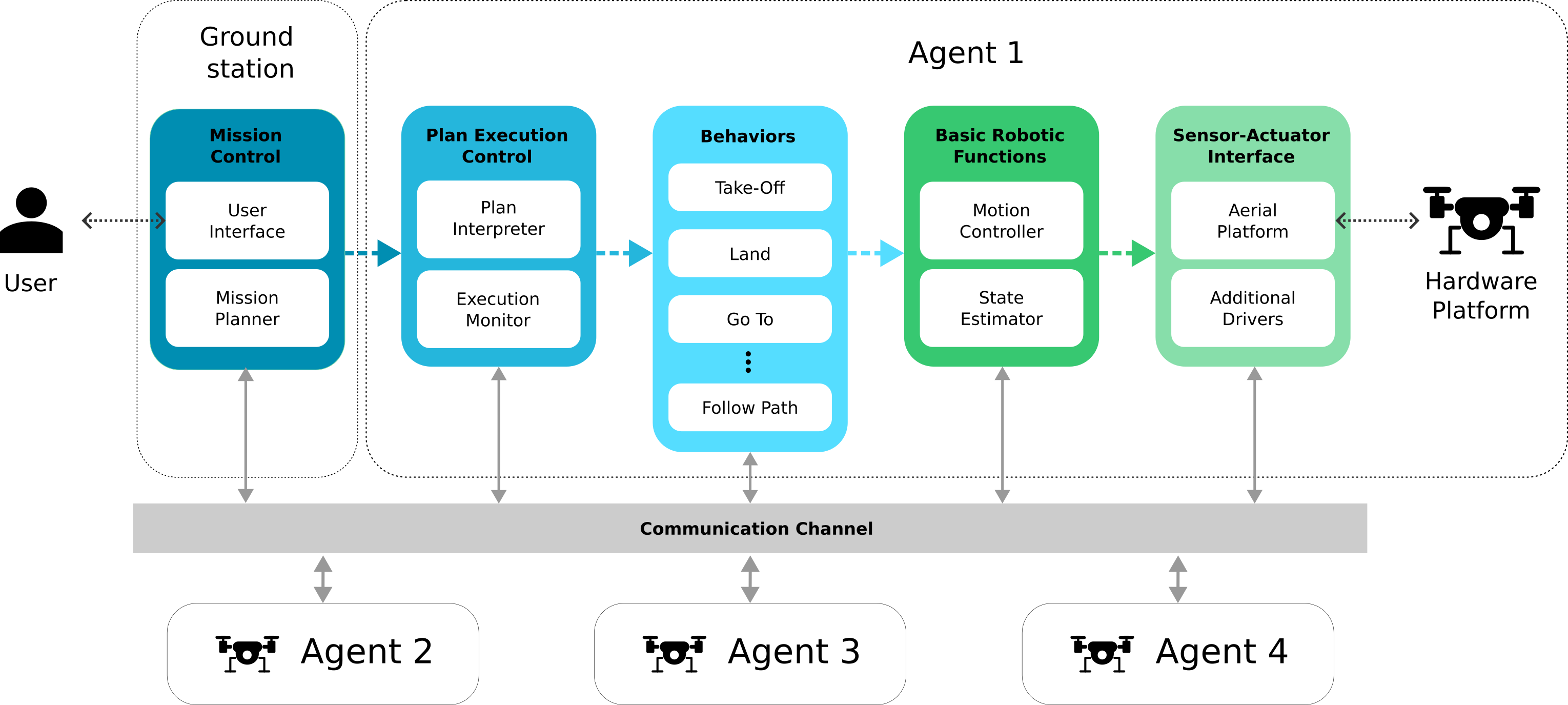}
\caption{Overview of the Aerostack2 framework architecture. Each colorized block represents a hierarchical layer, which can be constituted from multiple ROS 2 nodes or processes, represented as white blocks. All the nodes and agents are communicated through the ROS 2 communication channel, as is represented with grey arrows. The big colorized arrows going left to right, represent the \textit{command} flows, so that each blocks send commands to the right one. Nevertheless, each module can feed back meaningful information to the left modules through the communication channel, so the communication is bidirectional.}
\label{fig:software_stack}
\vspace{-0.5cm}
\end{figure*}

\begin{itemize}

\item \textbf{Sensor-Actuator interface}. These are components that serve as interfaces with multiple kinds of aerial platforms and sensors. Aerostack2 has defined standard interfaces that allow operating with both physical platforms and simulated platforms indistinctly. 
\item \textbf{Basic robotics functions}. This layer is dedicated to the components responsible for performing essential functions of aerial robotics, such as motion control or state estimation.
\item \textbf{Behaviors}. This layer in Aerostack2 implements specialized components called behaviors, which encapsulate specific robot skills for both reactive and goal-based actions, allowing self-monitoring and streamlined execution of tasks.
\item \textbf{Plan execution control}. 
This layer in Aerostack2 provides mechanisms for specifying and supervising mission plans, allowing developers to define complex tasks through the activation and deactivation of robot behaviors, thereby facilitating the autonomous execution of complex missions.
\item \textbf{Mission control}. The top layer of Aerostack2 offers user-friendly interfaces for mission planning, system monitoring, and debugging, tailored for both developers and non-developers.


\end{itemize}

In order to communicate all the components with each other, we have specified a common interface in the form of a set of message definitions for information exchange to facilitate process interoperability.

It is important to note that this modular organization of components is open to be used at any layer. This means that developers who use Aerostack2 may use the entire framework but also may use separately any intermediate layer or other individual component (e.g., the state estimator) for building a particular application. The following sections describe in more detail each layer of the software stack.

\subsection{\textbf{Sensor-Actuator Interface}}

One of the main objectives of the presented framework is to be able to implement autonomous aerial systems independently to the platform in which it will be finally deployed (considering that the sensory input and actuators are compatible).

To facilitate the implementation of different aerial platforms, Aerostack2 incorporates an \verb|AerialPlatform| abstract class responsible for managing the capabilities associated with the direct integration of various aerial platforms into the framework. This abstraction facilitates the integration of new platforms into the framework and ensures compatibility with the entire framework. The responsibility of this interface is to gather the sensory measurements from the aircraft and transmit them to the rest of the system. In addition, it is tasked with receiving actuator commands and other requests from the various layers of the Aerostack2 framework and relaying them to the aircraft in a platform-specific manner. 

These platforms ease the transition from simulation to real-world deployment because the logic modules remain agnostic to whether the system is operating on a real platform or in simulation. This also simplifies the implementation of heterogeneous aerial systems using different platforms.

\subsection{\textbf{Basic Robotic Functions}}

In aerial robotics, there are some functions that are vital for the basic operation of the system. In Aerostack2 we have called these Basic Robotic Functions, and each one has a component that is in charge of this functionality. Some of these functions are motion control and state estimation which support the autonomous operation of every aerial robot.

Due to the relevance of these components, we have found some general capabilities that these components are desired to implement:
\begin{enumerate}
    \item The implementation of the basic function has to be done in a plugin with a common interface.
    \item The plugins can be switched at run-time.
    \item The inputs and outputs can be adequate for standardizing the plugin interface.
\end{enumerate}




In this approach, each component is managed by a function manager, which is responsible for loading the plugins with each specific algorithm and managing how they interact with the rest of the framework. The plugin selector can also provide meta-control features, such as plugin replacement, whereas the input and output adapters adjust the input or output of the function plugin to the rest of the Aerostack2 framework. In the following, a more detailed explanation of the implementation of the Motion Controller and the State Estimator is provided.

\subsubsection{\textbf{Motion Control}} This component handles the motion references generated from the upper-level layers and convert them into actuator command signals which will be followed by the specific aerial platform. This component performs three tasks:
\begin{enumerate}
    \item Plugin Load and Selection: The module loads the different plugins that are required to operate. In this case, this plugin includes the implementation of different control laws to control the platform. The selection of the plugin can be done by the user or decided in runtime using predefined rules. When the motion reference can be handled directly by the aerial platform, the plugin can be bypassed.
    \item Control Mode Negotiation: Each plugin implements multiple combinations of input-output control mode. A control mode represents the set of different input signals that a controller plugin can handle. Similarly, each Aerial Platform has a set of control modes that are available to control it. The motion controller component is responsible for finding the best combination of modes that satisfies the requirements of the motion reference, the loaded plugin, and the aerial platform.
    \item Input-Output Adaptation: Based on the motion reference received (e.g. position, velocity, angle, etc.) in a specific reference frame, the module transforms this input into the desired coordinated frame of the loaded plugin. Similarly, some reference changes can be needed before sending the actuator commands to the aerial platform.
\end{enumerate}


\begin{figure}[!htb]
\centering
\includegraphics[width=0.45\textwidth]{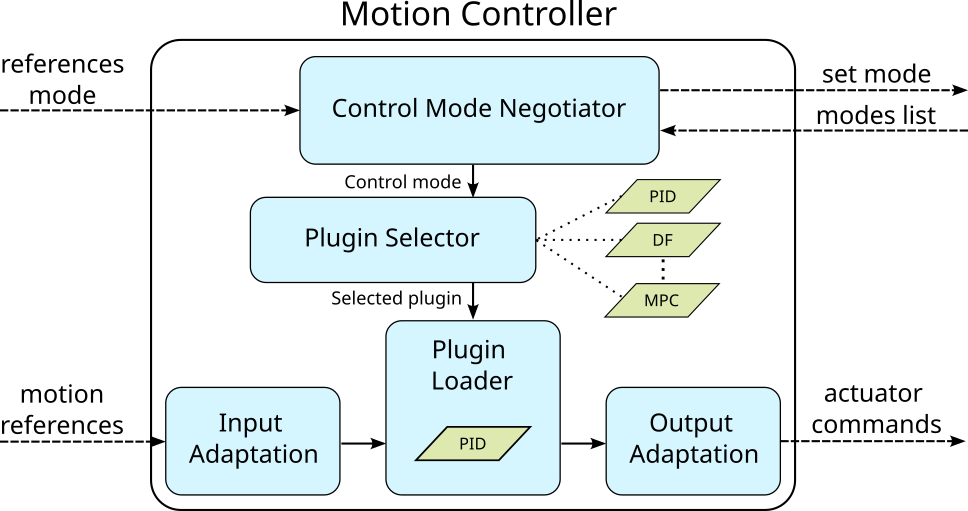}
\caption{Diagram of the motion controller component. Blue boxes represent the capabilities of it and how interact between them and the other modules. Plugins are represented with green trapezoids.}
\label{fig:motion_controller}
\end{figure}

\subsubsection{\textbf{State Estimator}} This module fuses the information received from different sensors to estimate the state of the robot, where the state refers to the position and speed of each aircraft over time. 
Similarly to the previous one, this module can load multiple state estimation algorithms in the form of plugins. The State Estimator module is responsible for:


\begin{enumerate}
    \item Plugin Load and Selection: The module shall load the different plugins that are required to operate. The selection of the plugin can be done by the user or decided in runtime using predefined rules.
    \item TF tree generation: The module is in charge of generating the transformation trees \cite{tully_foote_tf_2013} that will be used for the rest of the framework, allowing the system to represent information in different coordinate frames. This module is also in charge of managing the origin of the coordinated system in a multi-robot system.
    \item Geodesic to euclidean Conversion: When operating outdoors, this component is in charge of storing the geodesic operation origin, allowing the system to be able to work in the euclidean Space and translate geodetic coordinates. 
    \item Output Adaptation: The output information shall be adapted to be sent into the rest of the framework in a standardized way.
\end{enumerate}

\begin{figure}[H]
\begin{center}
\includegraphics[width=0.9\linewidth]{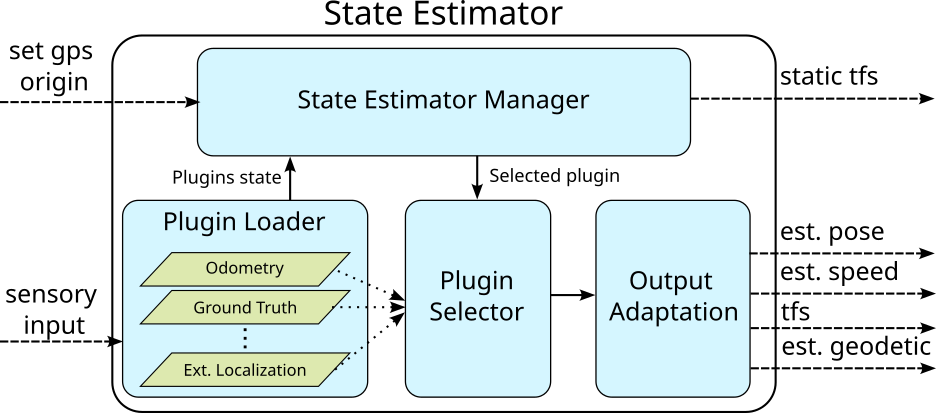}
\caption{Diagram of the State Estimator component. Blue boxes represent the capabilities of it and how interact between them and the other modules. Plugins are represented with green trapezoids.}
\label{fig:core_functions}
\end{center}
\vspace{-0.3cm}
\end{figure}

\subsection{\textbf{Behaviors}}

Aerostack2 uses a specialized type of component, called \textit{behavior}, which provides a logical layer to formulate mission plans in a uniform and simplified way, compared to the direct use of state estimators and actuator controllers. The notion of behavior in robotics has been used in the behavior-based paradigm \cite{Brooks1986} \cite{Arkin1998} \cite{Michaud2016}. In Aerostack2 each behavior corresponds to a specific robot skill related, for example, to flight motion, such as taking off, landing, hovering, and following a path, or other abilities (e.g., video recording, communication with other agents, generating a trajectory to reach a destination, etc.). 

\begin{figure}[t]
\begin{center}
\includegraphics[width=0.65\linewidth]{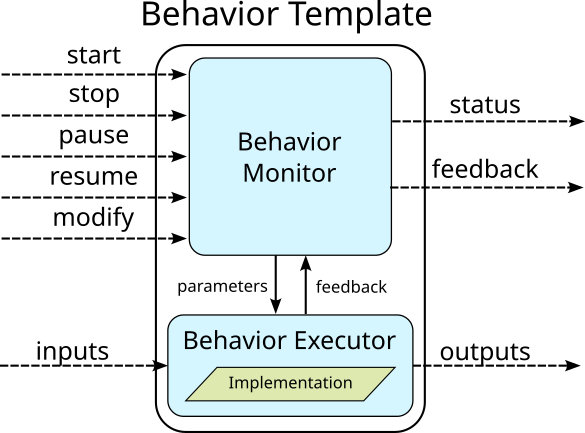}
\caption{Schematic representation of the behavior template architecture. The \textit{behavior monitor} processes external commands and monitors behavior execution by interfacing with the \textit{behavior executor}, which handles the underlying implementation. The system outputs include status and feedback, derived from the processed inputs.}
\label{fig:behavior}
\end{center}
\vspace{-0.5cm}
\end{figure}

Aerostack2 provides a behavior template architecture for programming behaviors that includes two distinct parts (Figure \ref{fig:behavior}). On the one hand, there is a \textit{behavior executor} that performs the execution using the algorithm used for each case. On the other hand, there is a \textit{behavior monitor} that interacts with the behavior executor to check its operation before, during, and after execution. The behavior monitor can adapt the execution by, for example, considering alternative modes of operation depending on the state of the environment\footnote{For example, in a landing maneuver, this can be done by descending until touching the ground, or it can be a landing maneuver that uses vision to center the drone on a landing spot. In both cases, the system calls the landing behavior in a similar way.} or response measures in the presence of faults. This feature adds adaptability and robustness to the architecture. The modular distribution of execution monitoring provided by Aerostack2 with behaviors also provides greater flexibility than architectures with centralized monitoring.

In addition, each software component that implements a behavior encapsulates the details of the algorithms used, providing a uniform interface that is common to all behaviors. This feature provides simplicity in describing mission plans. Using behaviors, a mission plan is expressed as a controlled sequence of activations (or deactivations) of multiple behaviors that may operate concurrently.  The result of each behavior execution is described in terms of success or failure, which is useful to determine the next step to be done during the mission.

In more detail, the uniform interface used to control the execution of the behavior uses the following basic services: \verb|start|, \verb|pause|, \verb|resume| and \verb|stop|. In addition, a service called \verb|modify| is used to change the parameters of a behavior without the need to stop its execution. The behavior interface informs about its execution state (e.g., idle, running, or paused) and periodic feedback about the concrete details on the running behavior. Aerostack2 behaviors have been implemented in a way that is fully consistent with standard ROS 2 actions.

\subsection{\textbf{Plan execution control}}
Aerostack2 provides several mechanisms to specify a mission plan and supervise its execution. The developer can write a plan to specify the set of tasks that a robot must perform on a particular mission. The specification of plans can be formulated in the form of activations and deactivations of different robot behaviors. In this way, the user can easily specify complex missions by combining tasks that the robot executes autonomously. Aerostack2 provides three main ways for generating and executing a plan: 

\subsubsection{Python API} Helps to interact with the Aerostack2 framework by implementing a \verb|DroneInterface| class that contains several methods to not only command the different drones, but also to retrieve meaningful information about their status. This is a convenient method that provides high flexibility for formulating plans with complex control regimes. Regarding the plan control, our API provides a set of functions to activate/deactivate behaviors or to command directly through motion reference messages. 

\subsubsection{Mission Interpreter} As a mission plan can be described as a sequence of behavior activation and deactivation, they can be described in a JSON format that can be directly translated into calls to the Python API by the \textit{Mission Interpreter}. This allows each drone to communicate the mission plan in a text format and then each drone will interpret it. This module supports conditional sentences that guide the mission execution.

\subsubsection{Behavior Trees} Finally one of the most used tools for describing plans in robotics is Behavior Trees \cite{Colledanchise_2018}. The modularity and hierarchy of behavior trees are useful not only during the mission plan design but also during mission execution thanks to graphical monitoring. Aerostack2 robot behaviors are compatible with BehaviorTreeCPP\footnote{\url{https://github.com/BehaviorTree/BehaviorTree.CPP/tree/v3.8}} library through their ROS 2 action interface.

\subsection{\textbf{Mission control and supervision}}

The top layer of the proposed architecture is mainly dedicated to the user interface. The components of this layer are designed to ease the definition of a mission to a human operator or to help with the supervision of the system. Regarding the level of user expertise, we can differentiate two blocks of tools: 

\subsubsection{Developers tools}
They show the internal status of the system, leading to a more detailed explanation of what is happening. 
The \textit{Alphanumeric Viewer} is a component that monitors the state of specific variables of the system, e.g. sensor measurements, values corresponding to state estimation, references for controllers, etc. The information is distributed in different panes to facilitate the search for a specific variable of the system.
On the other hand, tools like the \textit{Keyboard teleoperation} are useful to manipulate the drone swarm in a simple way, sending position and speed commands, which allows one to check the system behavior or take control when the autonomous logic fails. This tool allows to test and debug aerial systems fast and securely.

Both components operate in real-time through a Terminal User Interface (TUI), which makes these tools suitable for on-board debugging in low-power computers.

\subsubsection{End Users tools} In this case, the objective is to present all the information in a graphical and simple way so that non-developers can use and interact with the aerial system. In this category of components, Aerostack2 also provides a Graphical User Interface (GUI) to use the software framework through a web-based application. This tool is very convenient for facilitating rapid planning of a mission for one or several drones using graphical resources (e.g., a geographic map and graphical references), which can be done online or offline, allowing repeatability of missions, as shown in Fig.~\ref{fig:wgui_mission_planning}. This user interface is also useful to show graphically details of the mission execution, allowing its monitoring and modification in real-time.

%


\section{Experimental Validation}

This section aims to showcase the capabilities of the Aerostack2 to perform different autonomous drone missions in both the simulation and real world with different platforms.

\subsection{\textbf{Heterogeneous gate crossing}}

In this experiment, two different drones shall pass through two gates in a synchronized and cyclic fashion. The objective of this experiment is not only to test the simulation with real capabilities of the framework in an indoor environment but also to test the framework's platform independence. The experiment was carried out in two phases: The simulation phase, in which the mission is determined and validated using a simulator, and the real flights phase, in which the simulation results are compared with those obtained on real flights.
\subsubsection{Experimental Setup}
The simulation was performed using Gazebo simulator and the real experiment was carried out in an area of 60 $m^2$ equipped with a motion capture system (\textit{MoCap}) that provides the position of drones and gates within the capture area. The UAVs used for this experiment were two: a Bitcraze Crazyflie 2.1 and a Tello Ryze both with IR markers for mocap localization. Due to the limited payload of these micro-UAVs, the computation was performed by the ground station computer. 


\begin{table}[h]
    \centering
    \begin{tabular}{|Sc|c|c|}
        \hline
        \textbf{Module} & \textbf{Simulation} & \textbf{Real}\\
        \hline
        \hline
        Mission Control & \multicolumn{2}{c|}{Python API}\\
        \hline
        Behaviors & \multicolumn{2}{c|}{Motion, Platform, Trajectory Generator}\\
        \hline
        State Estimation & Ground Truth & Motion Capture System\\
        \hline
        Motion Control & \multicolumn{2}{c|}{DF controller}\\
        \hline
        Platforms & Gazebo Simulator & Crazyflie / Tello \\
        \hline
    \end{tabular}
    \caption{Comparison between Aerostack2 modules and plugins used in simulation and real flights of the crossing gates experiment.}
    \label{tab:exp:sim2real:comparison}
    \vspace{-0.1cm}
\end{table}

\subsubsection{Modules Used}
Table \ref{tab:exp:sim2real:comparison} shows a comparison between the components used in simulation and real experiments. For this mission, only basic behaviors are used: motion behaviors for taking off, following a path, and landing. The location of drones and gates is provided by the ground truth of the simulator or the \textit{mocap} system, and for control, a Differential Flatness (DF) based controller with trajectory references is used.

\subsubsection{Plan generation}
The mission was planned using the Aerostack2 Python API. In Listing \ref{lst:api} a simplified version of the mission with only one drone is shown. The synchronization mechanism is obtained by waiting for both drones to end their current behavior before activating the follow-up one. 

\begin{lstlisting}[language=Python, basicstyle=\scriptsize, label={lst:api}, caption=Simplified mission for Tello drone. The gates poses are provided by the \textit{MoCap} system.]]
...
n_laps=3
gate_names=["gate1","gate2"]
# Initialize drones interface
drone = DroneInterface("tello")
drone.arm()
drone.offboard()
drone.takeoff(height=2.0, speed=1.0)
for i in range(n_laps):
    for gate in gate_names: 
    #The path is on each gate coordinate system
        drone.follow_path([[-1, 0, 0],[1, 0, 0]],
            speed=1.0, 
            frame_id=gate)
drone.land(speed=0.5)
...
\end{lstlisting}

\subsubsection{Results}
\begin{figure}[htb]
\centering
\subcaptionbox{Simulation}
  [0.49\linewidth]{\includegraphics[height=4cm]{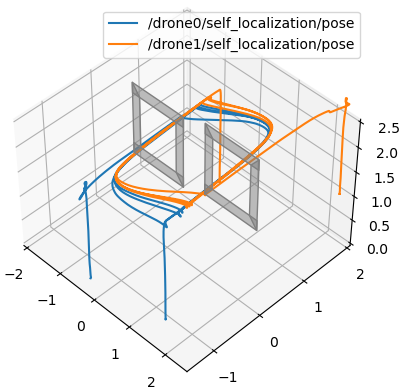}}
\subcaptionbox{Real}
  [0.49\linewidth]{\includegraphics[height=4cm]{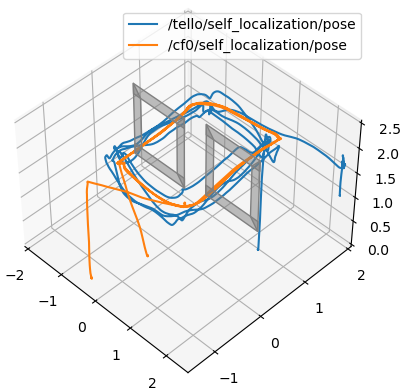}}
\caption{Comparison of the trajectories performed by the drones during the gate crossing experiments.}
\label{fig:exp:sim2real:trajectories}
\vspace{-0.3cm}
\end{figure}

\begin{figure}[H]
    \centering
    \includegraphics[width=0.9\linewidth]{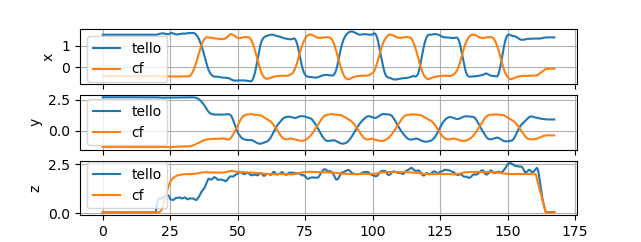}
    \caption{Plot of the evolution of the position on each axis of both drones during the real experiment.}
    \label{fig:position_graphics}
    \vspace{-0.3cm}
\end{figure}

In Fig. \ref{fig:exp:sim2real:trajectories} both simulation and real-world trajectories are plotted, showing the similarities in the trajectory followed in both environments. Moreover, in Fig. \ref{fig:position_graphics} it can be seen how both UAVs cross the gates in a synchronized and cyclical fashion.
The experiment proves that it is only needed to change the platform and the state estimation component (in this case, a plugin inside of it) to translate the experiment from simulation to the real world, even when the real system is heterogeneous. 




\subsection{\textbf{Cooperative area inspection}}

In this experiment, two high-end industrial drones will cooperate to inspect an area. In this scenario, a swarm of two drones is used to perform a back-and-forth aerial inspection.  As in the previous experiment, we tested the mission in a simulated environment before testing it in the real world, and a comparison was made between the two.

\subsubsection{Experimental Setup}
For simulation, we have used the DJI HITL simulator, which provides better dynamic modeling of the aerial platform. For both the simulation and the real-world experiments, we have used a DJI Matrice 300 and a DJI Matrice 350 both with a Nvidia Jetson AGX Xavier for onboard computing. Only the GUI was running on the ground station; once the mission had been loaded into each drone, all the computations had been done onboard each drone. In both real and simulated experiments, both drones shall inspect a 4500m$^2$ outdoors area.

\begin{table}[H]
    \centering
    \begin{tabular}{|Sc|c|c|}
        \hline
        \textbf{Module} & \textbf{Component}\\
        \hline
        \hline
        Mission Control &  Web GUI\\
        \hline
        Behaviors & Motion, Platform, Gimbal \\
        \hline
        State Estimation & GPS \\
        \hline
        Motion Control & PID controller\\
        \hline
        Platforms & DJI PSDK \\
        \hline
    \end{tabular}
    \caption{Aerostack2 modules and plugins used in both the HITL and real experiments for area coverage. Gimbal behaviors points the gimbal at a desired orientation.}
    \label{tab:exp:hetero:real}
    \vspace{-0.3cm}
\end{table}

\subsubsection{Modules Used}
Table \ref{tab:exp:hetero:real} shows the components used in both simulation and real experiments. For this mission, since we will use HITL simulation, all the modules remained the same in both experiments. In this case, the Web GUI is the component in charge of generating and uploading the mission that each drone is going to perform. We used the fused GPS signal for state estimation and a PID is used as the high-level controller.

\begin{figure}[H]
    \centering
    \includegraphics[width=0.45\textwidth]{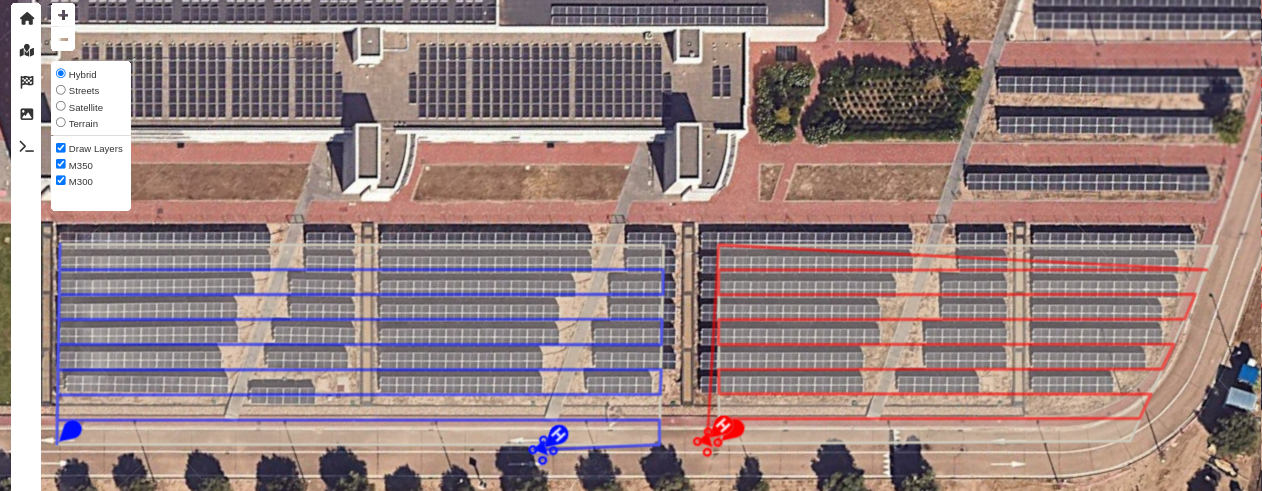}
    \includegraphics[width=0.45\textwidth]{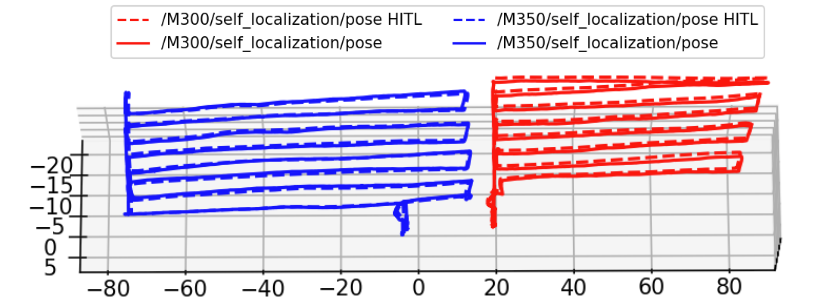}
    \caption{[Top] Aerostack2 GUI screenshot of the generated routes for both drones. [Bottom] Trajectories followed by both drones in the HITL simulation and the real flight.}
    \label{fig:wgui_mission_planning}
    \label{fig:out_exp:results}
    \vspace{-0.3cm}
\end{figure}

\subsubsection{Plan generation}
The mission was planned using our web GUI, which generated a path of GPS waypoints for both drones from a polygonal inspection area marked on a satellite image of the flight area. Once the mission was generated, both drones took off at the same time and each flew along the path, landing when finished. Figure \ref{fig:wgui_mission_planning} shows the mission generated for each drone. The GUI also allows for real-time monitoring of the operation.


\subsubsection{Results}

In Fig \ref{fig:out_exp:results}, can be shown that the trajectories followed by the two drones are so similar in both HITL simulation and the real world experiments and can be easily generated and replicated from the mission generated with the GUI, demonstrating the capabilities of Aerostack2 for working reliably with multiple drones in outdoor environments.

\subsection{\textbf{Aerostack2 for UAV Research}}
The Aerostack2 framework has been utilized to develop a wide range of aerial robotics systems within the research community. In \cite{rutinowski2023exploring}, the authors employed Aerostack2 to control a small UAV for reidentification of entities within an industrial setting. In \cite{mejias2024virtual}, a virtual Spring-Damper approach was developed and tested for UAV swarm formation in real-world environments. In \cite{luna2024multi}, the authors used Aerostack2 to test a novel approach for coverage path planning applications, incorporating replanning capabilities.

These are some examples that demonstrate the value of the presented framework to develop various applications and work towards developing novel algorithms.

\section{Conclusions and Future Work}


This paper presents a novel open-source framework designed for the development of aerial robotic systems, with a strong focus on multi-robot orientation, platform independence, versatility, and modularity. These features have been validated through a series of experiments in both simulated and real-world scenarios, demonstrating the effectiveness of the framework. Although our initial experiments involved only two drones simultaneously, the results confirm the framework's fundamental capabilities.


Future work will extend the testing of the framework to systems involving a larger fleet of drones, and explore its scalability and robustness in more complex scenarios. A key focus will be the development of an advanced communication component, which is essential for efficient information exchange between multiple vehicles. In addition, the plugin-oriented architecture will facilitate the development of meta-control capabilities, enabling further system enhancements. 

\bibliographystyle{unsrt}
\bibliography{scibib}

\end{document}